\documentclass[journal,twoside,web]{ieeecolor}
\usepackage[table,xcdraw]{xcolor}
\usepackage{tmi}
\usepackage{cite}
\usepackage{amsmath,amssymb,amsfonts}
\usepackage{algorithmic}
\usepackage{mathrsfs}
\usepackage{textcomp}
\usepackage{algorithmic}
\usepackage[caption=false]{subfig}
\usepackage{graphicx}
\usepackage{algorithm}
\usepackage[colorlinks=true,linkcolor=black,citecolor=blue,urlcolor=blue,]{hyperref}
\usepackage[doipre={doi:~}]{uri}

\usepackage{amssymb}
\usepackage{amsmath}
\usepackage{multirow}
\usepackage{booktabs}
\usepackage[misc]{ifsym}
\def\BibTeX{{\rm B\kern-.05em{\sc i\kern-.025em b}\kern-.08em
    T\kern-.1667em\lower.7ex\hbox{E}\kern-.125emX}}
\begin{document}
% \title{Iteratively Coupled Multiple Instance Learning with Confidence-based Embedder Fine-tuning for Whole Slide Image Classification}
\title{Rethinking Multiple Instance Learning for Whole Slide Image Classification: A Bag-Level Classifier is a Good Instance-Level Teacher}
\author{Hongyi~Wang,
        Luyang~Luo,~\IEEEmembership{Member,~IEEE},
        Fang~Wang,
        Ruofeng~Tong,
        Yen-Wei~Chen, ~\IEEEmembership{Member,~IEEE},
        Hongjie~Hu, 
        Lanfen~Lin,~\IEEEmembership{Member,~IEEE},
        and~Hao~Chen,~\IEEEmembership{Senior Member,~IEEE},
        % <-this % stops a space
\thanks{This work was supported by the National Key Research and Development Project (No. 2022YFC2504605), National Natural Science Foundation of China (No. 62202403), Hong Kong Innovation and Technology Fund (No. PRP/034/22FX), and the Project of Hetao Shenzhen-Hong Kong Science and Technology Innovation Cooperation Zone (HZQB-KCZYB-2020083). It was also supported in part by the Grant in Aid for Scientific Research from the Japanese Ministry for Education, Science, Culture and Sports (MEXT) under the Grant No. 20KK0234, 21H03470. (Corresponding authors: Lanfen Lin, Hao Chen.)}
\thanks{Hongyi Wang is with the College of Computer Science and Technology, Zhejiang University, Hangzhou 310063, China (e-mail: whongyi@zju.edu.cn).}
\thanks{Luyang Luo is with Department of Computer Science and Engineering, The Hong Kong University of Science and Technology, Hong Kong, China (e-mail: cseluyang@ust.hk).}
\thanks{Fang Wang is with Department of Radiology, Sir Run Run Shaw Hospital, Hangzhou 310016, China (e-mail: wangfang11@zju.edu.cn).}
\thanks{Ruofeng Tong is with College of Computer Science and Technology, Zhejiang University, Hangzhou 310063, China, and also with Research Center for Healthcare Data Science, Zhejiang Lab, Hangzhou 311121, China (e-mail: trf@zju.edu.cn).}
\thanks{Yen-Wei Chen is with College of Information Science and Engineering, Ritsumeikan University, Kusatsu 5250058, Japan, and also with College of Computer Science and Technology, Zhejiang University, Hangzhou 310063, China (e-mail: chen@is.ritsumei.ac.jp).}
\thanks{Hongjie Hu is with Department of Radiology, Sir Run Run Shaw Hospital, Hangzhou 310016, China (e-mail: hongjiehu@zju.edu.cn).}
\thanks{Lanfen Lin is with College of Computer Science and Technology, Zhejiang University, Hangzhou 310063, China (e-mail: llf@zju.edu.cn).}
\thanks{Hao Chen is with Department of Computer Science and Engineering, the Hong Kong University of Science and Technology, Hong Kong, China, and Department of Chemical and Biological Engineering, The Hong Kong University of Science and Technology, Hong Kong, China. He is also with HKUST Shenzhen-Hong Kong Collaborative Innovation Research Institute, Shenzhen, China. (e-mail: jhc@cse.ust.hk).}
}

\maketitle

\begin{abstract}
Multiple Instance Learning (MIL) has demonstrated promise in Whole Slide Image (WSI) classification. However, a major challenge persists due to the high computational cost associated with processing these gigapixel images. Existing methods generally adopt a two-stage approach, comprising a non-learnable feature embedding stage and a classifier training stage. Though it can greatly reduce the memory consumption by using a fixed feature embedder pre-trained on other domains, such scheme also results in a disparity between the two stages, leading to suboptimal classification accuracy. To address this issue, we propose that a bag-level classifier can be a good instance-level teacher. Based on this idea, we design Iteratively Coupled Multiple Instance Learning (ICMIL) to couple the embedder and the bag classifier at a low cost. ICMIL initially fix the patch embedder to train the bag classifier, followed by fixing the bag classifier to fine-tune the patch embedder. The refined embedder can then generate better representations in return, leading to a more accurate classifier for the next iteration. To realize more flexible and more effective embedder fine-tuning, we also introduce a teacher-student framework to efficiently distill the category knowledge in the bag classifier to help the instance-level embedder fine-tuning. Thorough experiments were conducted on four distinct datasets to validate the effectiveness of ICMIL. The experimental results consistently demonstrate that our method significantly improves the performance of existing MIL backbones, achieving state-of-the-art results. The code is available at: https://github.com/Dootmaan/ICMIL/tree/confidence\_based
\end{abstract}


\section{Introduction}
\label{sec:introduction}
\IEEEPARstart{W}{hole} Slide Images (WSIs), which are biomedical images scanned at the microscopic level, have become extensively utilized in cancer diagnosis \cite{niazi2019digital}. Compared to traditional microscope-based observation, WSIs offer advantages such as easier storage and analysis. However, their high-resolution nature presents challenges for computer-aided automatic analysis \cite{lu2021ai,su2022attention2majority}, as the currently limited GPU memory makes it infeasible to directly apply existing classification networks to these gigapixel images.
To tackle this issue, a widely adopted approach is the patch-based learning scheme \cite{cvpr2016em,zhang2018whole,chen2022deep}, which involves dividing a WSI into multiple small patches and training an end-to-end classification network exclusively on the patch level. Although the patch-level training consumes much less GPU memory, acquiring precise and comprehensive patch-level annotations for WSIs is prohibitively expensive due to their large size and resolution \cite{tizhoosh2018artificial}. 

\begin{figure*}[t]
\includegraphics[width=\textwidth]{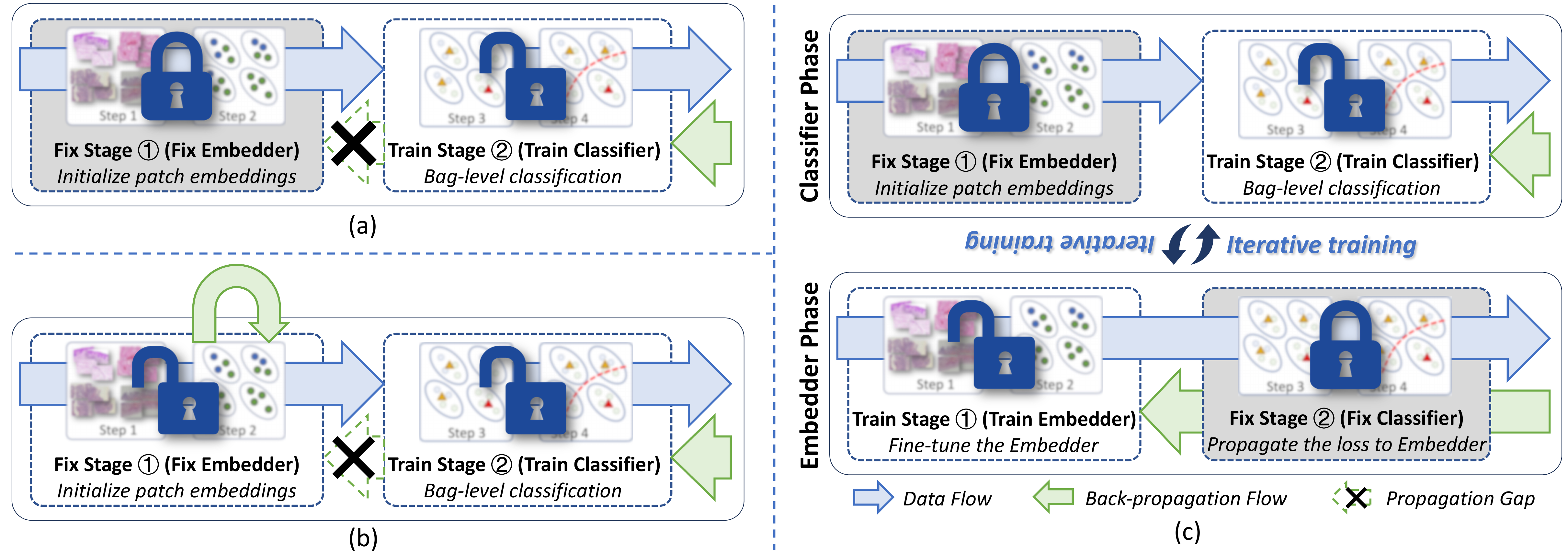}
\caption{Comparison between different MIL pipelines and our proposed ICMIL. (a) Ordinary MIL pipeline that uses a fixed patch embedder for patch feature extraction. There is a semantic gap between the embedder and the classifier in these methods because the embedder cannot be fine-tuned on the WSI data. (b) MIL pipelines that involve self-supervised fine-tuning of the patch embedder. Although the domain shift problem is mitigated in this way, there is still a semantic gap between the two stages of MIL since their training processes are separated and have no information exchange. (c) The proposed ICMIL pipeline, which consists of two training phases that can iteratively couple the two stages in MIL backbones. It solves the domain shift problem, and enables information exchange in the MIL backbones.} 
% \vspace{-3mm}
\label{MIL}
\end{figure*}

Therefore, MIL, as a weakly-supervised learning scheme that only requires slide-level labels, has become more and more widely used recently \cite{maron1997framework}. MIL methods treat each WSI as a bag, and consider the tiled non-overlapping patches within the WSI as the instances in the bag. A bag is classified as positive if at least one of its instances is positive, and as negative only if all instances are negative. To facilitate cost-effective training, the common pipeline for MIL-based WSI analysis typically consists of two stages as shown in Fig.~\ref{MIL}(a). The first stage is a non-learnable patch embedding stage, including tiling the gigapixel WSI into small non-overlapping patches and embedding them into feature vectors through a frozen patch embedder pre-trained on ImageNet \cite{imagenet}. The second stage is a trainable classification stage, comprising of aggregating the instance representations into bag-level representations and training a bag-level classifier on them. Although employing a frozen patch embedder reduces GPU memory consumption significantly, it also introduces a domain shift to the MIL classification model. Additionally, treating the embedding stage and the classification stage as two unrelated processes can also lead to lack of information exchange within the MIL pipeline, causing inconsistencies. Recent approaches have been proposed to fine-tune the patch embedder using task-agnostic methods, such as self-supervised techniques \cite{srinidhi2022self,dsmil,hipt} or weakly-supervised techniques \cite{liu2022multiple,wang2020ud,jin2023label}. However, as these two stages are still trained separately with different supervision signals, they lack joint optimization and still may cause inconsistencies in the MIL pipeline.

To address this issue, we propose that the bag-level classifier is a good instance-level teacher, which can be used to help in fine-tuning the embedder by guiding the generation of the instance pseudo-labels. Based on this idea, we develop ICMIL, a simple yet effective method that establishes a connection between the bag classifier and the patch embedder through an iterative optimization scheme. As shown in Fig.~\ref{MIL}(c), ICMIL consists of two phases. In the first phase (i.e., Classifier Phase), the patch embedder is frozen, and the subsequent aggregator and classifier are trained based on the instance representations generated by the embedder. In the second phase (i.e., Embedder Phase), the aggregator and the classifier are frozen, and the patch embedder is fine-tuned using instance-level pseudo-labels generated under the guidance of the bag-level classifier. To facilitate flexible and efficient embedder fine-tuning, we introduce a teacher-student-based method to enhance the knowledge distillation process from the bag-level classifier to the instance-level embedder. This method considers the bag-level classifier as the teacher of a hidden instance-level classifier, leveraging the category knowledge in the bag-level classifier to guide the instance-level embedder fine-tuning. 
When Embedder Phase is finished, ICMIL returns to Classifier Phase with the newly fine-tuned embedder, aiding in the training of a more accurate bag-level classifier, thereby improving the overall performance of the MIL backbone on the WSI dataset. If the desired model performance is not achieved, the iteration process can be repeated to ensure stronger coupling.
% ICMIL enables information exchange between the patch embedder and the bag-level classifier at a low cost, allowing the slide-level supervision to propagate to the embedder and facilitating more task-specific feature extraction.

The preliminary version of ICMIL \cite{icmil} was presented as a conference paper at the 26th International Conference on Medical Image Computing and Computer Assisted Intervention (MICCAI 2023). This extended work brings significant improvements, including proposing a new confidence-based embedder fine-tuning method for Embedder Phase, a novel bag augmentation method for Classifier Phase, and providing substantial extra experimental studies on four datasets. In summary, our contributions are: (1) We propose ICMIL, which can mitigate the domain shift and semantic gap problems in existing MIL backbones by iteratively coupling the bag classifier and the patch embedder in MIL methods during training. (2) We propose a confidence-based teacher-student approach to achieve effective and robust knowledge transfer from the bag-level classifier to the instance-level patch embedder, realizing weakly-supervised patch embedder fine-tuning. (3) We conduct thorough experiments with three representative MIL backbones on four datasets, and demonstrate the effectiveness of ICMIL.

% The rest of the paper is organized as follows: Section~\ref{section_related_works} provides an overview of related works and their limitations. Section~\ref{section_methodology} presents a detailed description of our proposed framework. Specifically, Section~\ref{section_general_idea} presents the general idea of ICMIL, while Section~\ref{section_phase1} and Section~\ref{section_phase2} provide detailed introductions to Classifer Phase and Embedder Phase of ICMIL, respectively. Section~\ref{section_experiments} reports the results of our extensive experimental validation and provides thorough analyses. Finally, Section~\ref{section_conclusion} summarizes and concludes our work.

% The rest of the paper is organized as follows. Related works and their limitations are given in Section~\ref{section_related_works}. The detailed description of our proposed framework is presented in Section~\ref{section_methodology}. Specifically, the general idea of ICMIL is presented in Section~\ref{section_general_idea}, and Classifier Phase and Embedder Phase of ICMIL are in detail introduced in Section~\ref{section_phase1} and \ref{section_phase2} respectively. Detailed experimental validation and results along with their thorough analyses are reported in Section~\ref{section_experiments}. Finally, our work is summarized and concluded in Section~\ref{section_conclusion}.

\section{Related Works}
\label{section_related_works}
\subsection{Multiple Instance Learning on WSI}
MIL has gained increasing popularity as an efficient solution for WSI classification, as it solely requires slide-level labels \cite{clam,dsmil} for training. In MIL methods, an ImageNet pre-trained feature embedder is typically employed to convert the $K$ patches within a WSI into $K$ feature vectors. This conversion significantly reduces the spatial dimensions from gigapixel to $K$$\times$$M$, where $M$ represents the dimensionality of the feature vectors. Subsequently, an aggregation module is utilized to fuse the $K$ feature vectors into a single 1$\times$$M$ feature vector, effectively representing the entire WSI. By training a bag-level classifier based on these feature vectors, the model can predict the category of the corresponding WSI at a low computational cost.

Max pooling and mean pooling are two widely used aggregation methods, but their simple mechanisms can also lead to sub-optimal performance. For instance, mean pooling has the potential to diminish the distinction between negative and positive instances, while max pooling may occasionally overlook the most discriminative patch-level instance. To address these limitations, Attention-based MIL (ABMIL) \cite{abmil} was proposed. ABMIL introduces a learnable aggregator that generates bag-level representations by utilizing attention scores assigned to the instance representations. Building upon this work, CLAM \cite{clam} was subsequently proposed, incorporating additional clustering branches for the bag-level classifier. Recent researches have also focused on leveraging the inter-instance information during the aggregation process. For instance, TransMIL \cite{shao2021transmil} proposed to use the self-attention mechanism \cite{transformer} to model the relationships between instances, and DSMIL \cite{dsmil} introduced a comprehensive multi-scale embedding fusion technique to generate more comprehensive patch representations.

\subsection{Embedder Fine-tuning in MIL}
In conventional MIL pipelines, the patch embedding stage and the classification stage are considered as separated and unrelated processes \cite{clam,abmil,rnnmil,shao2021transmil,dtfdmil}. When extracting features from gigapixel WSIs with limited GPU memory, it is common practice to bypass training the embedder and instead utilize a pre-trained model from ImageNet for feature extraction \cite{clam,abmil,rnnmil,shao2021transmil,dtfdmil}. However, this approach gives rise to a domain shift problem as the natural images in ImageNet can significantly differ from medical WSIs. Moreover, the two-stage training process introduces a semantic gap within the MIL pipeline, potentially compromising classification accuracy.
Presently, the most prevalent solutions to address domain shift challenges in MIL are task-agnostic fine-tuning methods based on Weakly-Supervised Learning (WSL) and Self-supervised Learning (SSL). In the ItS2CLR method \cite{liu2022multiple}, the authors propose generating instance-level pseudo-labels based on the aggregator attention score. Subsequently, instances with the top-k or bottom-k scores are considered positive or negative instances, respectively, for fine-tuning the embedder. Another approach, DSMIL \cite{dsmil}, employs self-supervised learning based on the SimCLR technique \cite{simclr}, enabling task-agnostic fine-tuning of the embedder. 
% Notably, the self-supervised pre-training, even with a simple ResNet-18 model as the patch embedder \cite{resnet}, yields good results.

However, despite these fine-tuning methods, the issue of broken loss back-propagation from the classification stage to the embedding stage persists in MIL pipelines. Consequently, task-specific fine-tuning of the patch embedder remains unrealized, leading to potential imperfections in optimizing the entire MIL pipeline. Figure~\ref{MIL} provides an intuitive comparison of different MIL pipeline designs. As depicted, conventional MIL pipelines overlook the domain shift problem and the semantic gap issue, relying on independent feature extraction, which significantly compromises model performance. On the other hand, MIL combined with WSL/SSL incorporates weak-supervision or self-supervision during patch embedder training, partially mitigating the domain shift problem. Nevertheless, since the fine-tuning process is separate from MIL training, it merely enables task-agnostic fine-tuning and fails to address the semantic gap between the embedder and the classifier in MIL. In contrast, our proposed method, ICMIL, effectively addresses this issue by establishing loss propagation from the classifier to the embedder, thus realizing task-specific fine-tuning of the embedder.

\begin{figure*}[t]
\center
\includegraphics[width=\textwidth]{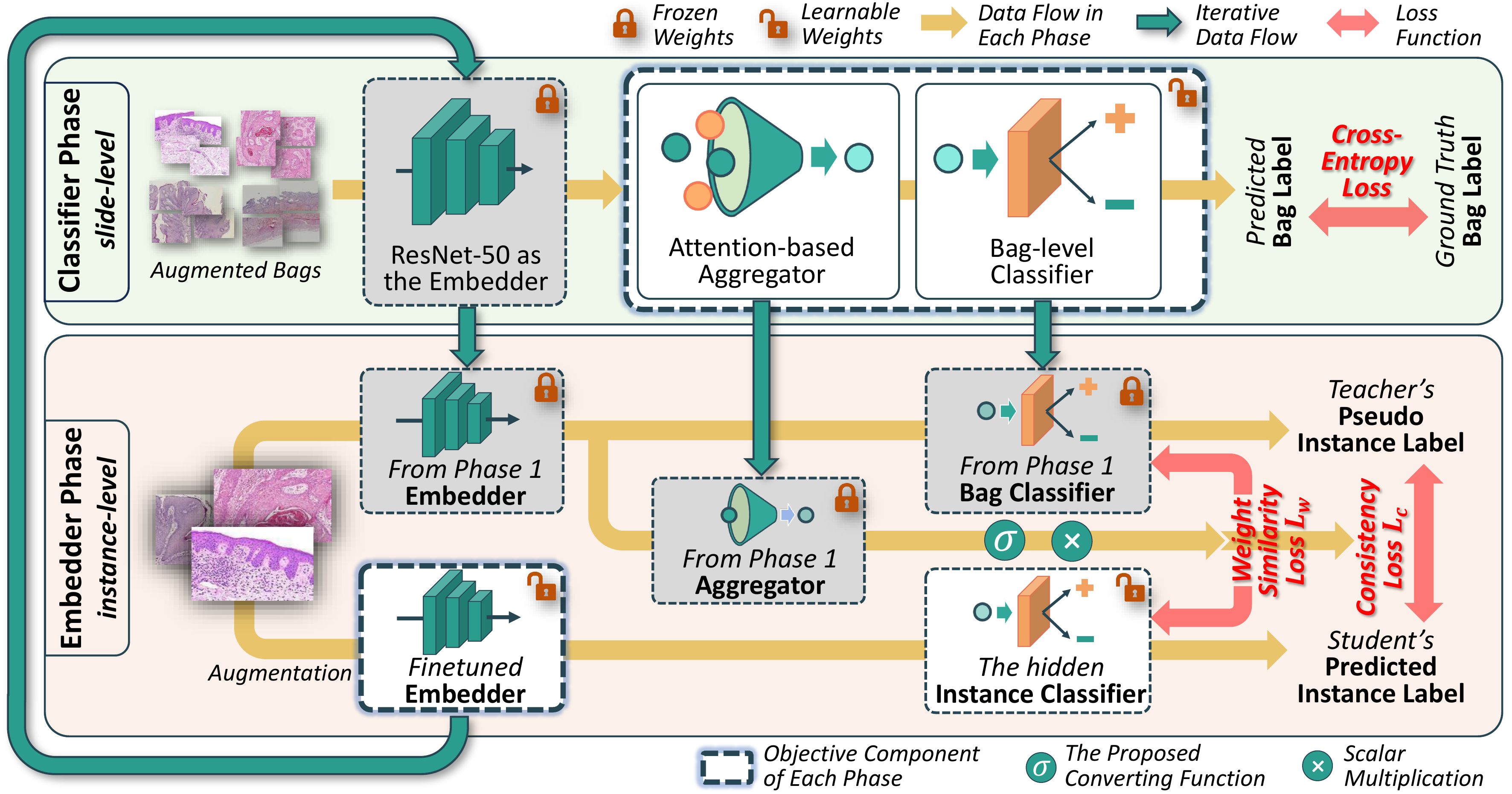}
\caption{A detailed view of the pipeline of ICMIL with confidence-based embedder fine-tuning method. The classic ABMIL is used as the ICMIL backbone for the example. } 
% \vspace{-3mm}
\label{ICMIL}
\end{figure*}

\section{Methodology}
\label{section_methodology}
\subsection{Overview of ICMIL}
\label{section_general_idea}
The fundamental concept of ICMIL is inspired by the classical machine learning technique, Expectation-Maximization (EM). EM is an iterative optimization method commonly used in problems involving hidden variables. Similarly, in MIL pipelines, we consider the instance representations obtained from the patch embedder as a hidden variable. Correspondingly in ICMIL, Classifier Phase optimizes the bag-level classifier based on the estimation of this hidden variable, and Embedder Phase fine-tune the embedder by maximizing the likelihood of the classifier's predictions, so the hidden variable can be re-estimated in the next iteration. 

% A schematic view of the proposed ICMIL is shown in Fig.~\ref{MIL}(c). The general idea of ICMIL is inspired by the classic machine learning optimization technique, EM. EM is an iterative optimization method mostly used in problems with a hidden variable. Similarly in MIL pipelines, we consider the instance representations from the patch embedder as a hidden variable. Therefore, the general pipeline should be: first, we fix this hidden variable with the estimated result from the patch embedder, and optimize the bag-level classifier; then, we fix the bag classifier and fine-tune the patch embedder to maximize the likelihood of classifier predictions. By doing so, Stage 1 and Stage 2 in existing MIL pipelines can be iteratively coupled, which indirectly bridges the loss propagation from the bag-level classifier to the patch embedder. 

To ensure the successful implementation of the fine-tuning scheme in Embedder Phase of ICMIL, it is essential for the bag-level representations and the instance-level representations to reside in the same latent space. This requirement is met by most MIL backbones, as their bag-level representations are derived from a linear combination of the instance representations. It is also advisable to use relatively simple aggregators in the MIL backbone to avoid introducing unnecessary complexity. Overly complex aggregators can lead to more significant disparities between the decision boundaries of the bag-level classifier and the hidden instance-level classifier. Consequently, this misalignment may introduce additional noise in the pseudo-labels during the fine-tuning of the embedder and slow down the convergence. 
% As we will use the bag-level classifier to guide the instance-level fine-tuning of the patch embedder in Embedder Phase of ICMIL, such a fine-tuning scheme would only be possible when the bag-level and instance-level representations stay in the same latent space. Generally, this prerequisite is satisfied by most existing MIL backbones, as their bag-level representations are all obtained from the linear combination of the instance representations. It is also recommended that the aggregators in the MIL backbone should avoid being over-complicated. Otherwise, it may lead to a large difference between the decision boundaries of the bag-level classifier and the hidden instance-level classifier, bringing more noise in the pseudo-labels during embedder fine-tuning.  

Different from previous works that solely utilize EM as a supportive tool for training either the patch embedder \cite{luo2020weakly, wang2017instance} or the bag-level classifier \cite{liu2022multiple}, we are the first to consider the entire MIL pipeline as an EM-like problem. It is important to note that while ICMIL draws inspiration from EM, its practical implementation incorporates unique designs that go beyond the traditional EM concept. For instance, in ICMIL, we perform two iterative optimization processes at two different scales, namely the slide level and the patch level. Furthermore, our optimization process differs from the traditional EM approach as we introduce a teacher-student-based framework during the instance-level embedder fine-tuning phase to mitigate the noise in the pseudo-labels. These modifications result in substantial differences in the mathematical expression and overall methodology of ICMIL compared to the original EM method, and to facilitate a better understanding, we will introduce the detailed pipeline of ICMIL based on the classic ABMIL backbone in the following paragraphs.
% To the best of our knowledge, different from previous works that only use EM as an assisting tool for aiding in the training of either the patch embedder \cite{luo2020weakly,wang2017instance} or the bag-level classifier \cite{liu2022multiple}, we are the first to consider the entire MIL pipeline as an EM-alike problem. It also should be noted that ICMIL is only inspired by EM for the main idea, but its practical implementation also includes some unique designs beyond the traditional EM concept. For example, the two iterative optimization processes in ICMIL are conducted at two different scales, namely slide-level and patch-level. Additionally, the optimization process is also different as we propose a teacher-student-based framework for the instance-level embedder fine-tuning phase to eliminate some noise in the pseudo-labels. These changes make the mathematical expression of ICMIL very different from the original EM method. To help better understand, we will introduce the detailed pipeline of ICMIL based on the classic ABMIL backbone in the following paragraphs. 

% The optimization goal can be mathematically presented as:

% \begin{equation}
%     \theta=\mathop{\mathrm{argmax}}\limits_{\theta}p(Y|\theta,X),
% \end{equation}

% \noindent where $\theta$ is the parameters of the entire MIL pipeline, which can be further divided into $\theta_1$ for the embedder and $\theta_2$ for the aggregator and the classifier; $Y$ is the true category of input $X$.

\subsection{Classifier Phase}
\label{section_phase1}

A schematic view of Classifier Phase in ICMIL is depicted in the upper section of Fig.~\ref{ICMIL}. In Classifier Phase, the primary objective is to train a bag-level classifier based on the fixed patch embedder. Since ICMIL is a framework that can suit different MIL methods, this phase can be implemented based on many existing MIL backbones, such as Max Pooling-based MIL, ABMIL, and DTFD-MIL. Nonetheless, to further improve the generalization ability of the MIL backbones, we also propose to use a pseudo-bag + bag mix-up scheme as data augmentation for this phase. 

\begin{figure}[t]
\center
\includegraphics[width=0.48\textwidth]{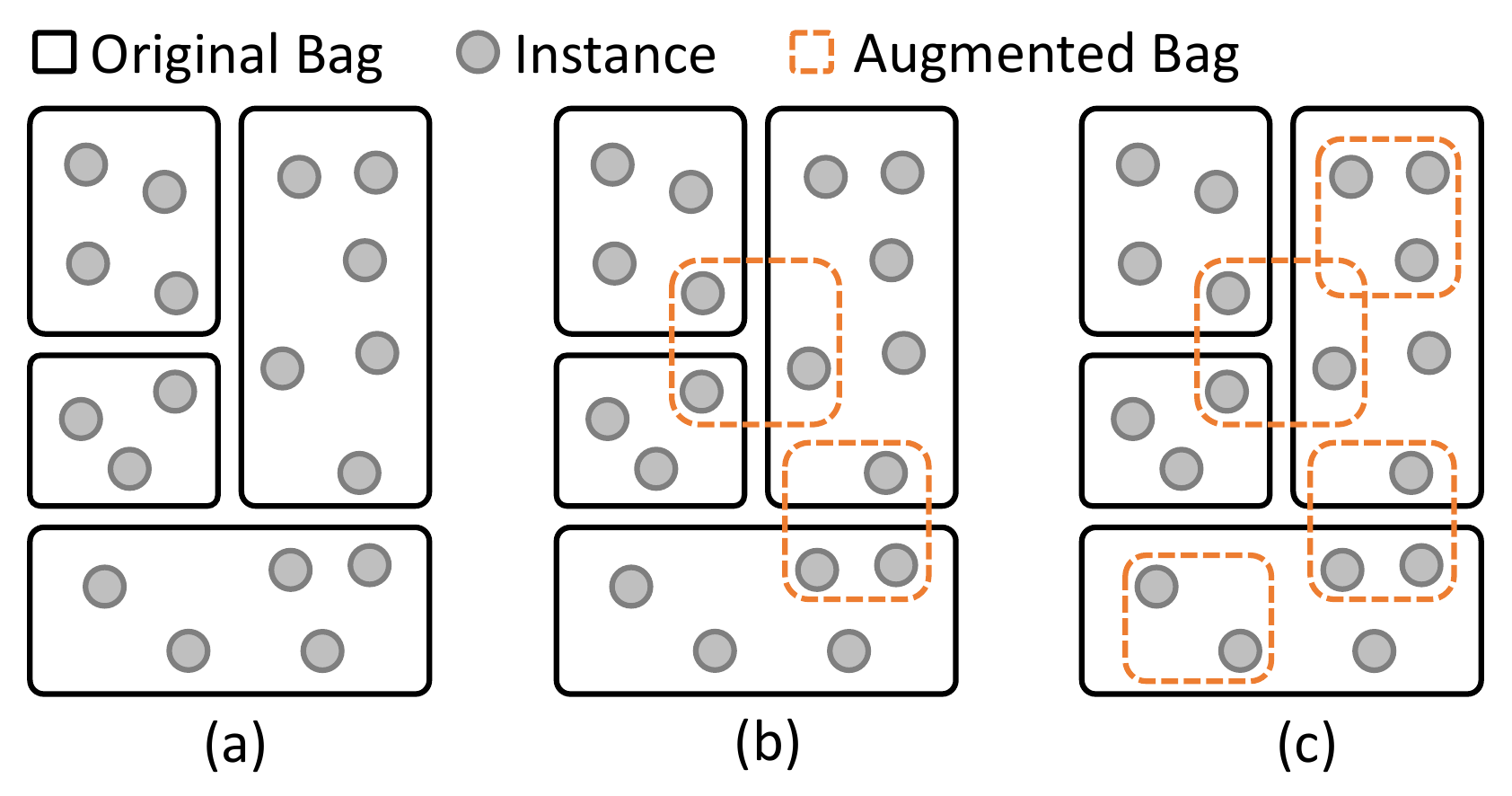}
\caption{Comparison of different augmentation methods. (a) The original dataset without augmentation. (b) Dataset after bag mix-up augmentation method. (c) Dataset after mix-up + pseudo-bag augmentation. } 
% \vspace{-3mm}
\label{augmentation}
\end{figure}

\subsubsection{Bag Augmentation}
Different from ordinary image augmentation methods on the instance level, we propose to use the more efficient bag-level augmentation on WSIs. Inspired from \cite{mixup, mixbag, pseudobagmixup}, we integrate the bag mix-up method with the pseudo-bag augmentation scheme, resulting in a higher generalization ability of the MIL model. 

% The detailed illustration of the augmentation method is shown in Fig.~\ref{augmentation}(c). Similar to the mix-up augmentation method \cite{mixup} used in natural image domain, bag mix-up is based on randomly selecting instances from several original bags to form a new bag. The corresponding label of the new bag is the weighted summation of the labels of the involved original bags. Additionally, the pseudo-bag augmentation method is also used in our pipeline. It is a simple method that generates a new bag by randomly sampling instances from one single original bag, and its label directly inherits that of the original bag as is assumed in many previous studies \cite{shao2021weakly,dtfdmil}. Such augmentation schemes can increase the diversity of the dataset, thereby enhancing the generalization ability of most MIL models. 

The mathematical expression of the augmentation method can be described as follows. Without loss of generality, let $(A,B)$ denote any pair of bags from the original dataset that $A\ne B$, and $y_A, y_B$ represent their labels. After partition, let $\{a_1, a_2, ..., a_n\}$ and $\{b_1, b_2, ..., b_n\}$ refer to their pseudo-bags respectively, where $n$ is a hyper-parameter set to 4 by default. We choose not to use a large $n$ because it can increase the possibility of a true-negative pseudo-bag inheriting the false-positive label. Then, we randomly mask $\lfloor \lambda \cdot n \rfloor$ bags in $A$ and $\lceil (1-\lambda) \cdot n \rceil$ in $B$ through a Beta distribution, i.e., $\lambda \sim Beta(\alpha,\alpha)$ and $\alpha$ is a non-negative hyper-parameter set to 1 by default. After that, the two masked bags are:

\begin{equation}
    A'=M_\lambda \odot \{a_1, a_2, ..., a_n\},
\end{equation}

\begin{equation}
    B'=(1-M_\lambda) \odot \{b_1, b_2, ..., b_n\},
\end{equation}

\noindent where $M_\lambda \in \{0,1\}^n$ is a binary mask indicating which pseudo-bag to mask and keep, and $\odot$ denotes the element-wise multiplication. With a possibility of $\gamma$, the proposed augmentation method directly returns $B'$ as an augmented pseudo-bag with its label directly being $y_B$. With a possibility of $1-\gamma$, the augmentation method returns the fusion of the two masked bags and uses $\lambda \cdot y_A + (1-\lambda)\cdot y_B$ as the label for this newly generated bag.

\subsubsection{Training the MIL Pipeline with a Frozen Embedder}
With the augmented bags, we then aim to train a bag-level classifier. In the case of ABMIL, we utilize the first three residual blocks of a ResNet-50 pretrained on ImageNet as the embedder, generating 1024-channel representation vectors for each patch. The subsequent aggregator and bag classifier can be directly optimized in the ordinary MIL way. 

Specifically in ABMIL, the aggregator employs an attention module to generate the attention scores for the instances, which are then used for the weighted summation of the instances to generate the bag-level representation for the WSI. Considering a WSI comprising $K$ patches, the mathematical expression of this process can be presented as:

\begin{equation}
    H=\sum_{k=1}^Ka_kh_k,
\end{equation}
\begin{equation}
    a_k=\frac{{\rm exp}\{\omega^T(tanh(V_1h_k)\odot sigm(V_2h_k))\}}{\sum_{j=1}^K{\rm exp}\{ \omega^T(tanh(V_1h_j)\odot sigm(V_2h_j))\}},
\end{equation}

\noindent where $H$ represents the bag-level representation, $a_k$ denotes the attention score for the $k$-th instance $h_k$ in the bag, and $K$ is the total number of instances. The matrices $\omega$, $V_1$, and $V_2$ are learnable parameters, and the symbols $\odot$, $\tanh(\cdot)$, and $\sigma(\cdot)$ refer to element-wise multiplication, hyperbolic tangent, and sigmoid activation functions, respectively. As shown, $H$ is obtained as a linear combination of the instance representations weighted by their corresponding attention scores. This ensures that it remains within the same latent space as the instance representations, satisfying the prerequisite of ICMIL. While the ABMIL has been used as an illustrative example here, similar analyses can be applied to other MIL backbones.
% Moreover, while the ABMIL framework has been used as an illustrative example here, similar analyses can be applied to other MIL backbones.

Next, the bag-level representation $H$ is forwarded to the bag-level classifier, which comprises a single linear layer that maps the representations to the predictions. The loss function employed in this phase is a standard cross-entropy loss, which is commonly utilized in existing MIL classification backbones. 

\subsection{Embedder Phase}
\label{section_phase2}
After the convergence of Classifier Phase, ICMIL proceeds to Embedder Phase to fine-tune the patch embedder using the guidance from the bag classifier. Since Embedder Phase operates at the patch level, it is crucial to determine appropriate pseudo-labels for the patch samples based on the components inherited from Classifier Phase. 

To realize a more flexible instance-level fine-tuning, we propose a teacher-student-based framework to distill the category knowledge in the bag classifier to the student branch for embedder fine-tuning. Further developed from the vanilla teacher-student method we previously presented in \cite{icmil}, we introduce confidence-based embedder fine-tuning in this work, whose structure is depicted in the lower part in Fig.~\ref{ICMIL}.

\subsubsection{Vanilla Teacher-Student Method}
\label{vanillafinetuning}

We start by giving a brief description about the vanilla teacher-student method. In the vanilla method, both the teacher and the student are composed of the patch embedder and the classifier inherited from Classifier Phase as shown in Fig.~\ref{ICMIL}. The teacher branch is set to be frozen to ensure stable guidance pseudo-label generation, while the student branch is set to be fully learnable conversely. The student embedder is set to learnable because we want to fine-tune it for the next round of iterative training; the student classifier is also set to learnable because it is designed to be the hidden instance-level classifier, whose decision boundary can be slightly different from the bag-level classifier. To make sure the learnable instance classifier would not be too different from the bag-level teacher, we also designed a weight similarity loss, which can be mathematically presented as:

\begin{equation}
    \mathscr{L}_w=\sum_{l=0}^L \sum_{c=0}^C \left[f(x)_c^l log\left( \frac{f(x)_c^l}{f'(x)_c^l}\right)\right],
\end{equation}

\noindent where $x$ is the input patch, $f(\cdot)$ represents the bag-level classifier in the teacher branch, $f'(\cdot)$ denotes the hidden instance-level classifier in the student branch, and $f(\cdot)_c^l$ indicates the $c$-th channel of $l$-th layer's output in $f(\cdot)$. 

Inspired by the self-distillation method proposed in \cite{noisystudent}, we introduce the noisy student branch to facilitate the transferability of knowledge from the teacher model to the student model. By exposing the student to augmented versions of the teacher's inputs, the student learns to generalize the teacher's knowledge across diverse variations of the input data, thereby empowering the MIL backbone with heightened robustness and generalization capabilities. Based on this design, there is also a consistency loss $\mathscr{L}_c$ for the teacher's and student's output. This is a conventional loss function for teacher-student frameworks to make sure that the student branch can learn the distilled knowledge from the teacher's guidance and make similar predictions. $\mathscr{L}_c$ can be mathematically presented as:

\begin{equation}
    \mathscr{L}_c=\sum_{c=1}^{C} \left[f(x)_c log\left( \frac{f(x)_c}{f'(x')_c}\right)\right],
\end{equation}

\noindent where $x'$ is the input patch $x$ after augmentation.

% Finally, the overall loss function $\mathscr{L}$ for this phase can be written as:
% \begin{equation}
%     \mathscr{L}=\mathscr{L}_c+\alpha \mathscr{L}_w.
% \end{equation}

% while $\alpha$ is a hyper-parameter adjusting the balance between these two terms. It is set to 0.5 empirically in our experiments.

\subsubsection{Confidence-based Teacher-Student Method}
\label{confidencefinetuning}

% \begin{figure}[t]
% \includegraphics[width=0.48\textwidth]{ICMIL_confidence_based.pdf}
% \caption{Pipeline of the proposed confidence fine-tuning method. Compared with the vanilla teacher-student method, this confidence-based framework further introduces the aggregator from Classifier Phase for evaluating the importance of each sample. } 
% % \vspace{-3mm}
% \label{ICMIL_confidence_based}
% \end{figure}

Although the vanilla teacher-student method already achieves a good performance, it treats all instances equally and does not distinguish between pseudo-labels of different uncertainty. To address this limitation, we propose a confidence-based teacher-student fine-tuning method in this work, which leverages the aggregator trained in Classifier Phase to evaluate the confidence of each instance's pseudo-label. 

% Reweighting and resampling are two widely used strategies in various machine learning tasks, including model boosting, semi-/weakly-supervised learning, and reinforcement learning \cite{liu2020mesa, schlegel2019importance}. Theoretically, these two strategies tend to have similar performance since they are statistically equivalent \cite{an2020resampling}. However, in practical implementation, there are still some minor differences between them, such as stability, convergence speed, and model performance. These differences are mainly influenced by the dataset, model structure, and optimization algorithm. When considering these components as an integrated system, resampling and reweighting can exhibit different advantages, such as generating more diverse samples \cite{kang2019few} or correcting sampling bias with stochastic gradients \cite{an2020resampling}.
% In the vanilla teacher-student fine-tuning method, we use the trained bag-level classifier from Classifier Phase as the teacher of the hidden instance-level classifier, and realize flexible fine-tuning for the patch embedder. However, this method treats all the instances the same, and cannot distinguish the more discriminative instances from the less significant ones. Therefore, we propose a confidence-based teacher-student fine-tuning method for Embedder Phase of ICMIL, of which the practical implementation can be further categorized into reweighting the instances and resampling the instances. The complete pipeline of the confidence-based fine-tuning method with resampling strategy is shown in Fig.~\ref{ICMIL}.

\begin{figure}[t]
\center
\includegraphics[width=0.4\textwidth]{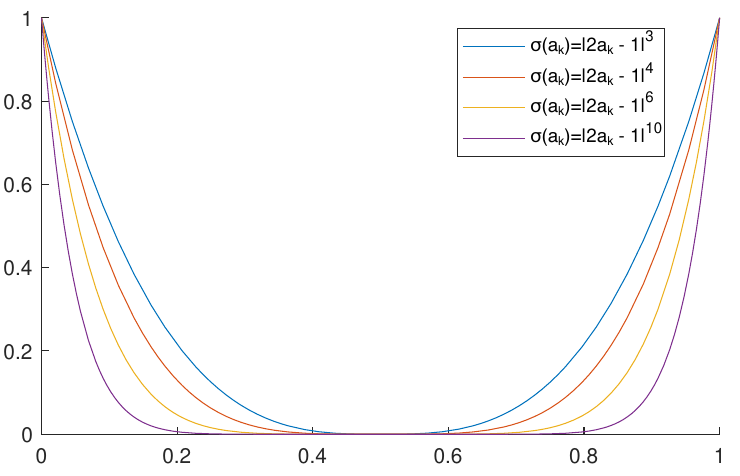}
\caption{The proposed converting layer for converting the attention scores into confidence scores. } 
% \vspace{-3mm}
\label{activation_function}
\end{figure}

% \begin{algorithm}[t]
% \caption{Pseudo Code for Embedder Phase of ICMIL Training Pipeline w/ Confidence-based fine-tuning Method}
% \begin{algorithmic}
% \STATE \textbf{Input: WSI dataset X, MIL components including embedder $g(\cdot)$, aggregator $a(\cdot)$ trained in Classifier Phase and bag classifier $f(\cdot)$ trained in Classifier Phase}
% % \STATE \textbf{Initialize: Initialize Embedder $g(\cdot)$ with ImageNet pre-trained ResNet50}
% \STATE \textbf{Output: Optimal MIL model $f(a(g(\cdot)))$}

% \STATE \# Embedder Phase: fix $f(\cdot)$ and fine-tune the embedder $g(\cdot)$
% \STATE $ f'(\cdot)\gets f(\cdot).copy() $
% \STATE $ g'(\cdot)\gets g(\cdot).copy() $
% \STATE $ f(\cdot).require\_grad\gets \textbf{False} $
% \STATE $ g(\cdot).require\_grad\gets \textbf{False} $
% \STATE $ \textbf{for } x_i \in X: $
% \STATE \hspace{0.5cm}$ \textbf{ for } x_i^j \in Tiled(x_i): $
% \STATE \hspace{1.0cm}$ y_{tch}\gets f(g(x_i^j)) $
% \STATE \hspace{1.0cm}$ y_{stu}\gets f'(g'(Augment(x_i^j))) $
% \STATE \hspace{1.0cm}$ loss_2\gets L_c(y_{stu},y_{tch})+0.5*L_w(f'(\cdot),f(\cdot)) $
% \STATE \hspace{1.0cm}$ w=\sigma(a(x_i^j))$
% % \STATE \hspace{1.0cm}$ \textbf{if } resampling=\textbf{True}:$
% % \STATE \hspace{1.5cm}$ w=B(w)$\quad \#$B(\cdot)$ is the Bernoulli mask
% \STATE \hspace{1.0cm}$ loss_2 \gets loss_2*w $
% \STATE \hspace{1.0cm}$ loss_2.backward() $
% \STATE \hspace{1.0cm}$ optimizer_2.step() $
% \STATE $ g(\cdot)\gets g'(\cdot) $
% \end{algorithmic}
% \label{alg2}
% \end{algorithm}

% \textbf{Confidence-based Reweighting.}
As shown in Fig.~\ref{ICMIL}, in this confidence-based teacher-student fine-tuning method, we employ the trained aggregator in Classifier Phase to generate scores for the input instances. However, in MIL, attention scores often primarily emphasize positive instances and tend to disregard negative ones. Yet, for the embedder fine-tuning task, both positive and negative samples are required. Consequently, directly utilizing the attention score of each instance as its confidence score would result in a severe bias. To address this issue, we identify instances with high attention scores as high-confidence positive instances, and consider instances with extremely low attention scores as high-confidence negative samples. Therefore, we introduce a specialized converting layer $\sigma$ for the attention scores, as depicted in Fig.~\ref{activation_function}. This converting layer effectively projects the attention scores into confidence scores, which can be subsequently used to adjust the weights in the loss functions. The mathematical expression of the converting layer is:

\begin{equation}
    \sigma(a_k)=|2\cdot a_k-1|^\beta,
\end{equation}

\noindent where $a_k$ is the attention score for the $k$-th instance that ranges from 0 to 1 after min-max normalization, and $\beta$ is the hyper-parameter adjusting the smoothness of the converting function. $\beta$ is set to 6 by default in our experiments.

Altogether, the loss function for this confidence-based fine-tuning method is:

\begin{equation}
    \mathscr{L}=\sigma(a(x))\cdot(\mathscr{L}_c+\alpha \mathscr{L}_w),
\end{equation}

\noindent where $x$ is the input patch image as defined before, and $a(x)$ is its corresponding attention score generated by the aggregator.

% \begin{figure*}[t]
% \center
% \includegraphics[width=0.95\textwidth]{HCC_pipeline.pdf}
% \caption{The collection pipeline of the HCC dataset used in our experiments. } 
% % \vspace{-3mm}
% \label{hcc_pipeline}
% \end{figure*}

% \textbf{Confidence-based Resampling.} 
% Different from reweighting, the resampling strategy additionally employs a Bernoulli distribution mask for selecting high-confidence samples. This Bernoulli distribution $B(p)$ considers the confidence score of each instance as its parameter $p$, controlling the sampling ratio of each instance. Consequently, each instance has a possibility of $p$ of being sampled during the embedder fine-tuning phase, or a possibility of $1-p$ of being ignored. 
% In other words, the output of this Bournoulli mask $B(p)$ could be mathematically described as:

% \begin{equation}
%     B(\sigma(a(x))|p_x)=\left\{
%     \begin{aligned}
%     1, if\ \sigma(a(x))> p_x\\
%     0, if\ \sigma(a(x))< p_x
%     \end{aligned}
%     \right.
% \end{equation}

% In this way, the loss function for the resampling-based fine-tuning method is:

% \begin{equation}
%     \mathscr{L}=B(\sigma(a(x)))\cdot(\mathscr{L}_c+\alpha \mathscr{L}_w).
% \end{equation}

% \noindent where $\sigma(a(x))$ serves as the parameter $p$ for $B(\cdot)$.

% \textcolor{red}{(maybe this section should be removed because i feel that it does not provide much useful information and makes our pipeline become even more complicated.)}

\section{Experiments}
\label{section_experiments}
\subsection{Datasets and Metrics}
\subsubsection{Datasets}
There are four datasets used in our experiments. The first one is the widely used public benchmark Camelyon16 \cite{luo_breast1,luo_breast2,camelyon16}. It consists of 270 training WSIs and provides an official test split with 129 cases. The positive WSIs in this dataset exhibit a severe class imbalance problem, with typically only about 10\% tumor patches. Although detailed patch-level annotations are provided for this dataset, they are not used in our experiments but only used for visualization.

The second dataset is the public TCGA-Lung Dataset (i.e., TCGA Non-Small Cell Lung Carcinoma Dataset), which is composed of samples from the TCGA-LUAD (Lung Adenocarcinoma) project with 534 cases and the TCGA-LUSC (Lung Squamous Cell Carcinoma) project with 512 WSIs.

The third dataset is the public TCGA-RCC (Renal Cell Carcinoma) Dataset. This dataset contains three sub-type projects, namely Kidney Chromophobe Renal Cell Carcinoma (TGCA-KICH, 121  slides), Kidney Renal Clear Cell Carcinoma (TCGA-KIRC, 519 slides), and Kidney Renal Papillary Cell Carcinoma (TCGA-KIRP, 300 slides). 

The last dataset is a private hepatocellular carcinoma (HCC) dataset collected from Sir Run Run Shaw Hospital, Hangzhou, China. This dataset comprises 1140 tumor WSIs scanned at 40$\times$ magnification. The task is to grade each case based on the Edmondson-Steiner (ES) grading scores. The ground truth labels are provided by experienced pathologists, where cases with ES grading I and II are labeled as low-risk cases, and cases with ES grading III and IV are considered as high-risk cases.

\subsubsection{Metrics}

For all the datasets, we evaluate the performance of different methods with three metrics, namely Area Under Curve (AUC), F1-score (F1), and Accuracy (Acc). The curve used for calculating AUC is the receiver operator characteristics (ROC) curve.

\begin{table}[t]
\caption{Comparison between different embedder fine-tuning methods on Camelyon16 (w/o bag augmentation). }
\center
\begin{tabular}{@{}cc|ccc@{}}
\toprule
Backbone               & Method     & \begin{tabular}[c]{@{}c@{}}AUC\\(\%)\end{tabular}      & \begin{tabular}[c]{@{}c@{}}F1\\(\%)\end{tabular}      & \begin{tabular}[c]{@{}c@{}}Acc\\(\%)\end{tabular}   \\ \midrule
\multirow{3}{*}{Max Pooling} & Naive Pseudo-label & 84.0 & 73.5 & 80.8  \\ &  Vanilla ICMIL  & 85.2 & 74.4 & 81.9 \\
                          & Confi-based ICMIL & 85.2 & 74.4 & 81.9  \\ \midrule
\multirow{3}{*}{ABMIL}  & Naive Pseudo-label & 88.5 & 78.8 & 83.9 \\ & Vanilla ICMIL  & 90.0 & 80.5 & 86.6 \\
                          & Confi-based ICMIL & 91.8 & 82.5 & 86.6 \\ \midrule
\multirow{3}{*}{DTFD-MIL} & Naive Pseudo-label & 93.1 & 84.9 &  90.0 \\ & Vanilla ICMIL  & 93.7 & 87.0 & 90.6 \\
                          & Confi-based ICMIL & 94.7 & 90.5 & 92.9 \\ \bottomrule
\end{tabular}
\label{tab_ablation}
\end{table}

\begin{table}[t]
\renewcommand\tabcolsep{4.5pt}
\caption{Parameter study on $\beta$ of the converting layer for confidence-based ICMIL method (w/o bag augmentation). }
\center
\begin{tabular}{@{}ccccccccc@{}}
\toprule
\multirow{2}{*}{\begin{tabular}[c]{@{}c@{}}Backbone of\\ Confi-based ICMIL\end{tabular}} & \multicolumn{4}{c}{Camelyon16}    & \multicolumn{4}{c}{HCC}           \\ \cmidrule(l){2-5} \cmidrule(l){6-9} 
  & $\beta$=2 & $\beta$=4 & $\beta$=6 & $\beta$=8 & $\beta$=2 & $\beta$=4 & $\beta$=6 & $\beta$=8 \\ \midrule
ABMIL  & 90.9   & 91.3   & 91.8   & 91.7   & 87.3   & 87.9   & 87.9   & 87.4   \\
DTFD-MIL  & 94.7   & 94.4   & 93.9   & 94.0   & 88.1   & 88.1   & 87.4   & 87.2   \\ \bottomrule
\end{tabular}
\label{parameter_study}
\end{table}

\begin{table*}[t]
\caption{Comparision of different bag-level augmentation strategies on Camelyon16 dataset with three different MIL backbones. The results are acquired with the confidence-based ICMIL method. }
\center
\begin{tabular}{@{}cccccccccccccc@{}}
\toprule
\multirow{3}{*}{Method}    & \multirow{3}{*}{\begin{tabular}[c]{@{}c@{}}Augmentation \\ Method\end{tabular}} & \multicolumn{3}{c}{Camelyon16} & \multicolumn{3}{c}{HCC} & \multicolumn{3}{c}{TCGA-Lung} & \multicolumn{3}{c}{TCGA-RCC} \\ \cmidrule(l){3-5} \cmidrule(l){6-8} \cmidrule(l){9-11}  \cmidrule(l){12-14} 
                             &                                      & \begin{tabular}[c]{@{}c@{}}AUC\\(\%)\end{tabular}      & \begin{tabular}[c]{@{}c@{}}F1\\(\%)\end{tabular}      & \begin{tabular}[c]{@{}c@{}}Acc\\(\%)\end{tabular}    & \begin{tabular}[c]{@{}c@{}}AUC\\(\%)\end{tabular}      & \begin{tabular}[c]{@{}c@{}}F1\\(\%)\end{tabular}      & \begin{tabular}[c]{@{}c@{}}Acc\\(\%)\end{tabular}   & \begin{tabular}[c]{@{}c@{}}AUC\\(\%)\end{tabular}      & \begin{tabular}[c]{@{}c@{}}F1\\(\%)\end{tabular}      & \begin{tabular}[c]{@{}c@{}}Acc\\(\%)\end{tabular}    & \begin{tabular}[c]{@{}c@{}}AUC\\(\%)\end{tabular}      & \begin{tabular}[c]{@{}c@{}}F1\\(\%)\end{tabular}      & \begin{tabular}[c]{@{}c@{}}Acc\\(\%)\end{tabular}     \\ \midrule
\multirow{4}{*}{\begin{tabular}[c]{@{}c@{}}Confidence-\\based ICMIL\\(w/ MaxPooling)\end{tabular}} & None                      & 85.2          & 74.4        & 81.9         &   86.6     & 87.3      & 82.0       & 96.3         & 90.4        &  90.1   & 97.0 & 84.7 & 88.0     \\
& Pseudo-bag                      & 86.7          & 77.7        & 82.3         &   86.8     & 87.7      & 81.5       & 95.9         & 90.0        &  89.9   & 97.2 & 84.8 & 88.2     \\
  & Mix-up   &  87.0         & 78.8        &  83.1        &  86.6      & 88.0      & 82.1       &   96.3       &  90.7       &  90.7    & 97.2 & 84.5 & 88.3    \\
  & Both (Ours)     &  87.4         &  79.6       &  84.3        & 87.3       & 88.1      & 82.5       &  96.6        & 90.9        &  90.7   & 97.5 & 85.1 & 88.5     \\ \midrule
\multirow{4}{*}{\begin{tabular}[c]{@{}c@{}}Confidence-\\based ICMIL\\(w/ ABMIL)\end{tabular}}       & None     &  91.8         &   82.5      &    86.6      & 87.9       &  89.0     &  84.1      &  96.3        & 90.7        & 90.2     & 97.8 & 86.4 & 89.2     \\
& Pseudo-bag   &  91.7         &   82.8      &    87.1      & 88.3       &  89.1     &  84.2  &  96.3         & 90.4        &  90.1   & 97.5 & 86.1 & 89.2     \\
  & Mix-up   &   92.2   &  83.7  &  87.4    & 88.5       & 89.0      & 83.9       &  95.3        &  90.0       &  89.7    & 97.9 & 86.1 & 88.9    \\
   & Both (Ours)  & 92.4    & 81.9   &  88.2    & 88.5      & 89.9       &  84.7        &  96.6       &  90.3   & 89.9   & 98.0 & 86.9 & 89.7  \\ \midrule
\multirow{4}{*}{\begin{tabular}[c]{@{}c@{}}Confidence-\\based ICMIL\\(w/ DTFD-MIL)\end{tabular}}       & None                      &  94.7         &  90.5      &   92.9    & 88.1     &  89.7     &  84.0      &  97.2       & 91.4        & 91.3     & 97.6 & 88.4 & 89.8   \\
& Pseudo-bag                    &  94.1         &  90.0      &   92.2    & 88.0     &  89.5     &  83.6    & 96.3         & 90.4        &  90.1  & 97.1 & 88.1 & 89.5      \\
                             & Mix-up   &  94.7   & 89.7  &  92.2    & 88.3      & 88.5      &  83.1      &  96.7        & 91.9        & 90.9   &  97.9 &  88.3 & 89.9    \\
                             & Both (Ours) & 95.0    & 88.2    & 91.3  &  88.9   & 90.2      & 84.5       & 96.8         &    91.9     &  91.3  & 97.9 & 88.6 & 90.2      \\ \bottomrule
\end{tabular}
\label{augmentation_comparison}
\end{table*}

\begin{figure*}[t]
\center
\subfloat[Results w/ Max Pooling-based MIL]{\includegraphics[width = 0.32\textwidth]{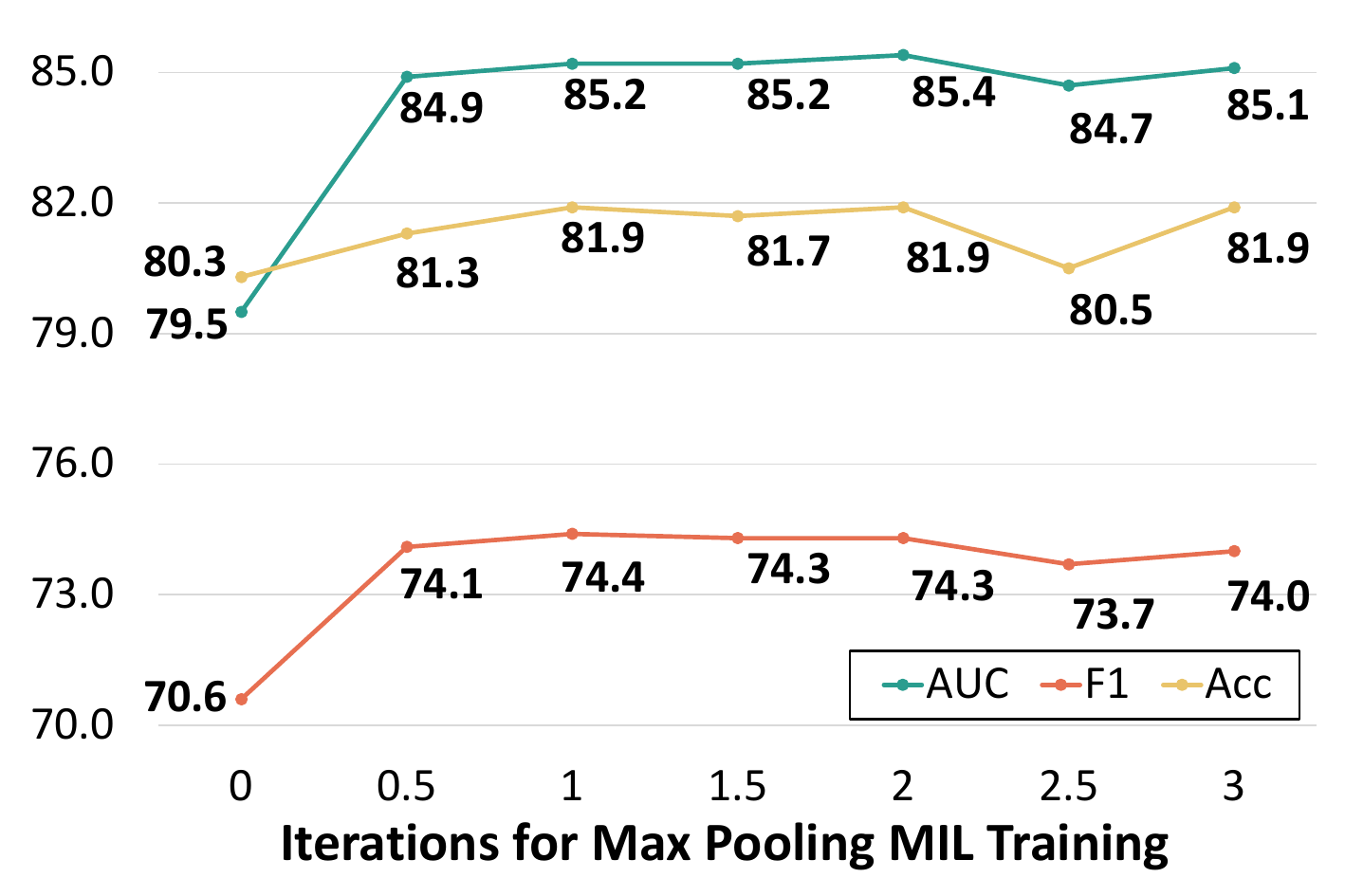}}
	\hfill
\subfloat[Results w/ ABMIL]{\includegraphics[width = 0.32\textwidth]{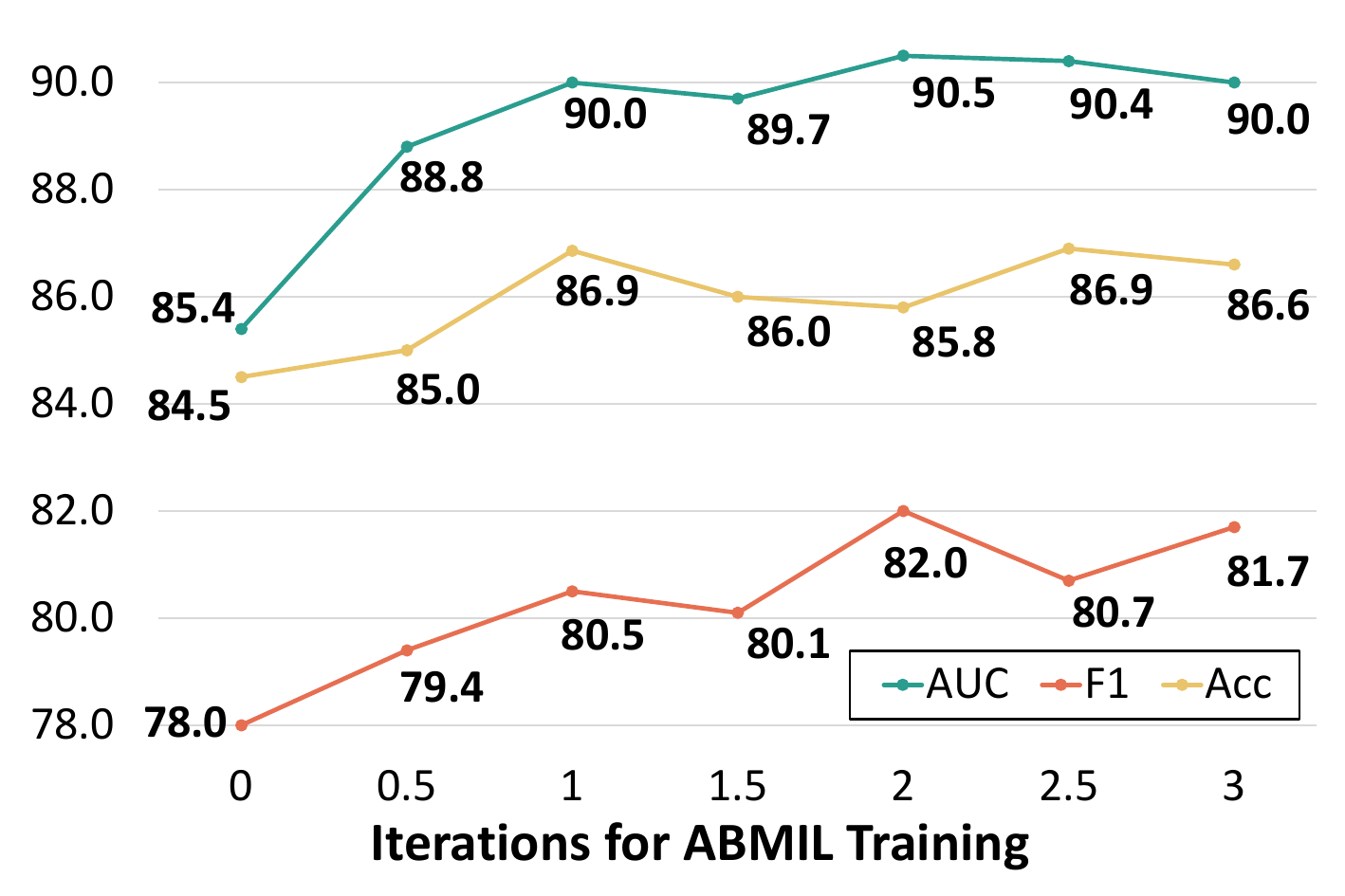}}
	\hfill
\subfloat[Results w/ DTFD-MIL]{\includegraphics[width = 0.32\textwidth]{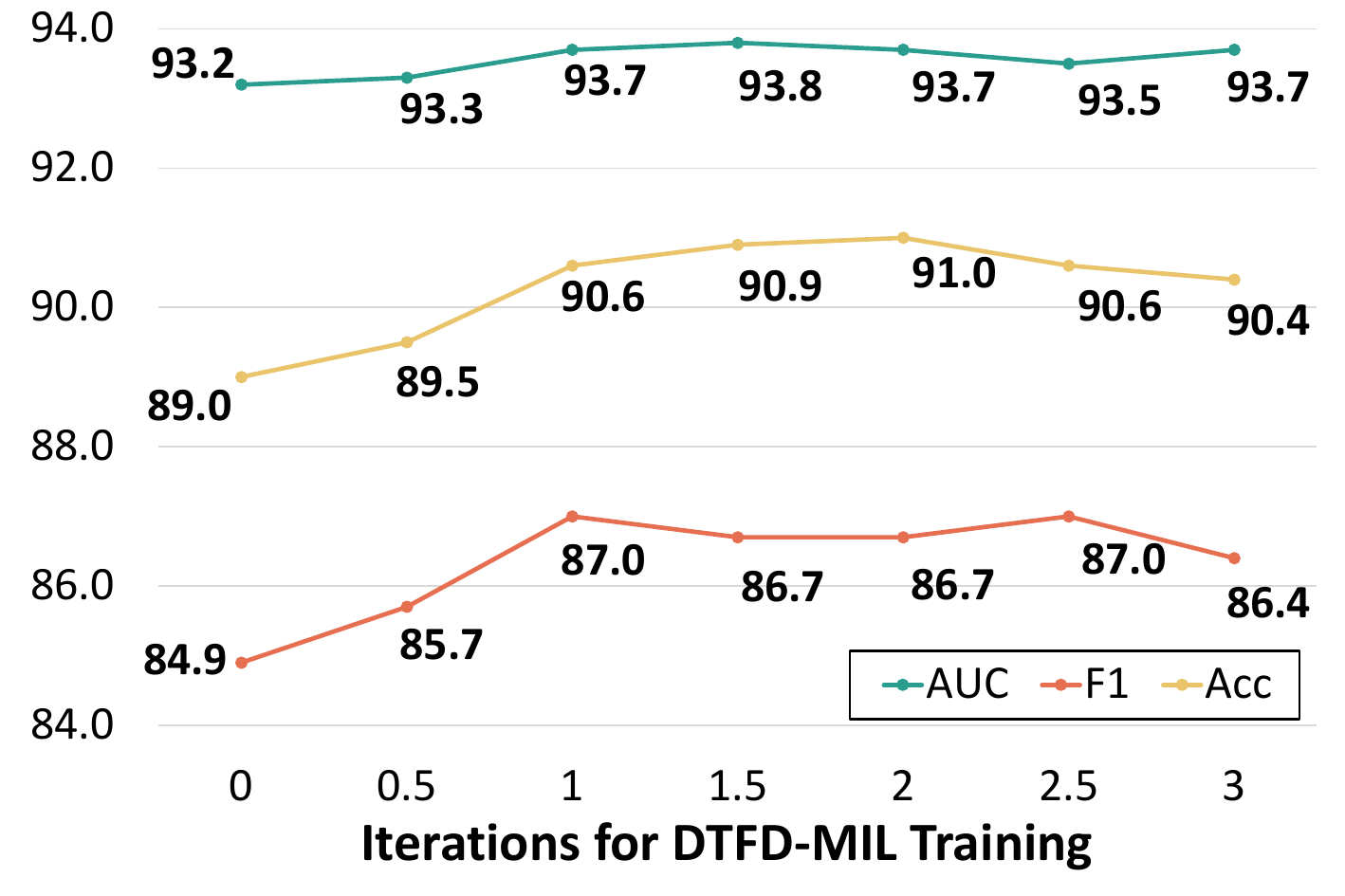}}
\caption{AUC performance of ICMIL on Camelyon16 with different ceasing iterations for training. } 
% \vspace{-3mm}
\label{ablation_study_iteration}
\end{figure*}

\subsection{Implementation Details}
For the Camelyon16 and the two TCGA datasets, we tile the WSIs into non-overlapping patches of size 256$\times$256 at the 20$\times$ magnification level, following the same settings as the previous state-of-the-art method DTFD-MIL \cite{dtfdmil}. On the other hand, for the HCC dataset, we tile the WSIs into 384$\times$384 patches at the 40$\times$ magnification level based on the advice of pathologists. As a result, the Camelyon16 dataset yields approximately 3.7 million patches, the TCGA-Lung dataset generates around 15.3 million patches, the TCGA-RCC dataset generates around 12.9 million patches, and the HCC dataset produces approximately 17.4 million patches.

To comprehensively evaluate the proposed ICMIL method, we select three MIL approaches as the backbones for ICMIL, namely Max Pooling-based MIL, ABMIL, and the state-of-the-art method DTFD-MIL. During Classifier Phase, we utilize a learning rate of 2e-4 for the Adam optimizer \cite{adam} and train the MIL backbone for 200 epochs. For Embedder Phase, we use a learning rate of 1e-5 for the Adam optimizer \cite{adam}, while the batch size is set to 80 for the HCC dataset and 100 for the others. The Camelyon16 results are reported on the official test split, while other datasets adopt a 7:1:2 split for training, validation, and testing. All experiments are implemented with PyTorch on a machine with two Xeon E5-2640 v4 CPUs, 96GB RAM, and an Nvidia Tesla M40 (12GB).

\begin{table*}[t]
\renewcommand\tabcolsep{5.5pt}
\renewcommand\arraystretch{1.2}
\caption{Comparison with other methods on the four datasets. We use the confidence-based fine-tuning method to get the results, with $\beta$=2 for DTFD-MIL and $\beta$=6 for the others. The best results are in bold, while the second-best ones are underlined.}
\center
\label{experimental_results}
\begin{tabular}{ccccccccccccc}
\hline
\multirow{3}{*}{Method}                         & \multicolumn{3}{c}{Camelyon16} & \multicolumn{3}{c}{HCC} & \multicolumn{3}{c}{TCGA-Lung} & \multicolumn{3}{c}{TCGA-RCC}  \\ \cmidrule(l){2-4} \cmidrule(l){5-7}  \cmidrule(l){8-10} \cmidrule(l){11-13} 
                        &  \begin{tabular}[c]{@{}c@{}}AUC\\(\%)\end{tabular}      & \begin{tabular}[c]{@{}c@{}}F1\\(\%)\end{tabular}      & \begin{tabular}[c]{@{}c@{}}Acc\\(\%)\end{tabular}       & \begin{tabular}[c]{@{}c@{}}AUC\\(\%)\end{tabular}      & \begin{tabular}[c]{@{}c@{}}F1\\(\%)\end{tabular}      & \begin{tabular}[c]{@{}c@{}}Acc\\(\%)\end{tabular}     & \begin{tabular}[c]{@{}c@{}}AUC\\(\%)\end{tabular}      & \begin{tabular}[c]{@{}c@{}}F1\\(\%)\end{tabular}      & \begin{tabular}[c]{@{}c@{}}Acc\\(\%)\end{tabular}    & \begin{tabular}[c]{@{}c@{}}AUC\\(\%)\end{tabular}      & \begin{tabular}[c]{@{}c@{}}F1\\(\%)\end{tabular}      & \begin{tabular}[c]{@{}c@{}}Acc\\(\%)\end{tabular}   \\ \hline
Mean Pooling-based MIL            & 60.3    & 44.1    & 70.1    & 76.4  & 83.1  & 73.7 & 89.5 & 83.5 & 83.5 & 95.9 & 82.8 & 86.2\\
Max Pooling-based MIL  & 79.5    & 70.6    & 80.3    &  80.1      &  84.3      & 76.8  & 95.9 &  90.2 & 89.6 & 96.6 & 84.1 & 87.5\\
RNN-MIL \cite{rnnmil}    & 87.5    & 79.8    & 84.4    & 79.4       &  84.1      & 75.5  & 89.4 & 83.1 & 84.5  & 96.2 & 83.1 & 86.6  \\
ABMIL \cite{abmil} & 85.4    & 78.0    & 84.5    & 81.2  & 86.0  & 78.1 & 95.2 & 89.8 & 89.7 & 97.3 & 86.2 & 89.0 \\
DSMIL \cite{dsmil}   & 89.9    & 81.5    & 85.6    & 86.1       & 86.6       & 81.4 & 93.9 & 87.6 & 88.8  & 97.2 & 85.0 & 88.9   \\
CLAM-SB  \cite{clam}   & 87.1    & 77.5    & 83.7    &  82.1      & 84.3       & 77.1   & 94.4 & 86.4 & 87.5 & 97.0 & 84.6 & 87.8  \\
CLAM-MB \cite{clam} & 87.8    & 77.4    & 82.3    &  81.7      & 83.7       & 76.3 & 94.9 & 87.4 & 87.8  & 97.5 & 87.1 & 89.5   \\
TransMIL \cite{shao2021transmil}  & 90.6    & 79.7    & 85.8    & 81.2       & 84.4       & 76.7 & 94.9 & 87.6 & 88.3  & 97.2 & 85.0 & 87.6   \\
DTFD-MIL \cite{dtfdmil}  & 93.2    & 84.9    & 89.0    & 83.0  & 85.5  & 78.1 & 96.0 & 90.9 & 90.7 & 97.6 & 88.4 & 89.8 \\ \hline
\vspace{-0.5mm}$\rm \mathop{Vanilla\ ICMIL}\limits_{(w/\ Max\ Pooling)}$ \cite{icmil} &  $\mathop{85.2}\limits_{(+5.7)}$      & $\mathop{74.7}\limits_{(+4.1)}$         & $\mathop{81.9}\limits_{(+1.6)}$    & $\mathop{86.6}\limits_{(+6.5)}$  & $\mathop{87.3}\limits_{(+3.0)}$     & $\mathop{82.0}\limits_{(+5.2)}$ & $\mathop{96.3}\limits_{(+0.4)}$  & $\mathop{90.4}\limits_{(+0.2)}$   & $\mathop{90.1}\limits_{(+0.5)}$ & $\mathop{97.0}\limits_{(+0.4)}$  & $\mathop{84.7}\limits_{(+0.6)}$   & $\mathop{88.0}\limits_{(+0.5)}$ \\
\vspace{-0.5mm}$\rm \mathop{Vanilla\ ICMIL}\limits_{(w/\ ABMIL)}$ \cite{icmil}   & $\mathop{90.0}\limits_{(+4.6)}$   & $\mathop{80.5}\limits_{(+2.5)}$    & $\mathop{86.6}\limits_{(+2.1)}$  & $\mathop{87.1}\limits_{(+5.9)}$        &  $\mathop{88.3}\limits_{(+2.3)}$       & $\mathop{83.3}\limits_{(+5.2)}$  & $\mathop{95.9}\limits_{(+0.7)}$  &  $\mathop{90.4}\limits_{(+0.6)}$  &  $\mathop{90.0}\limits_{(+0.3)}$   & $\mathop{97.6}\limits_{(+0.3)}$  &  $\mathop{86.5}\limits_{(+0.3)}$  &  $\mathop{89.1}\limits_{(+0.1)}$    \\
\vspace{-0.5mm}$\rm \mathop{Vanilla\ ICMIL}\limits_{(w/\ DTFD\text{-}MIL)}$ \cite{icmil}  &  $\mathop{93.7}\limits_{\underline{(+0.5)}}$      & $\mathop{87.0}\limits_{\underline{(+2.1)}}$   & $\mathop{90.6}\limits_{\underline{(+1.6)}}$  & $\mathop{87.7}\limits_{(+4.7)}$     &  $\mathop{89.1}\limits_{(+3.6)}$      &  $\mathop{83.5}\limits_{(+5.4)}$  & $\mathop{96.7}\limits_{\underline{(+0.7)}}$   &  $\mathop{91.1}\limits_{\underline{(+0.2)}}$  & $\mathop{91.0}\limits_{\underline{(+0.3)}}$  & $\mathop{97.7}\limits_{(+0.1)}$  &  $\mathop{88.4}\limits_{\underline{(+0.0)}}$  &  $\mathop{89.8}\limits_{\underline{(+0.0)}}$    \\ \hline
\vspace{-0.5mm}$\rm \mathop{Confi\text{-}based\ ICMIL}\limits_{(w/\ Max\ Pooling)}$ &  $\mathop{87.4}\limits_{(+7.9)}$      & $\mathop{79.6}\limits_{(+9.0)}$         & $\mathop{84.3}\limits_{(+4.0)}$    & $\mathop{87.3}\limits_{(+7.2)}$  & $\mathop{88.1}\limits_{(+3.8)}$     & $\mathop{82.5}\limits_{(+5.7)}$  & $\mathop{96.6}\limits_{(+0.7)}$ & $\mathop{90.9}\limits_{(+0.7)}$ & $\mathop{90.7}\limits_{(+1.1)}$  & $\mathop{97.5}\limits_{(+0.9)}$  &  $\mathop{85.1}\limits_{(+1.0)}$  &  $\mathop{88.5}\limits_{(+1.0)}$    \\
\vspace{-0.5mm}$\rm \mathop{Confi\text{-}based\ ICMIL}\limits_{(w/\ ABMIL)}$    & $\mathop{92.4}\limits_{(+7.0)}$   & $\mathop{81.9}\limits_{(+3.9)}$    & $\mathop{88.2}\limits_{(+3.7)}$  & $\mathop{88.5}\limits_{\underline{(+7.4)}}$        &  $\mathop{89.9}\limits_{\underline{(+3.9)}}$       & $\mathop{84.7}\limits_{\underline{(+6.6)}}$     & $\mathop{96.6}\limits_{(+1.4)}$        &  $\mathop{90.3}\limits_{(+0.5)}$       & $\mathop{89.9}\limits_{(+0.2)}$    & $\boldsymbol{\mathop{98.0}\limits_{(+0.7)}}$  &  $\mathop{86.9}\limits_{(+0.7)}$  &  $\mathop{89.7}\limits_{(+0.7)}$      \\
\vspace{-0.5mm}$\rm \mathop{Confi\text{-}based\ ICMIL}\limits_{(w/\ DTFD\text{-}MIL)}$ 
&  $\boldsymbol{\mathop{95.0}\limits_{(+1.8)}}$      & $\boldsymbol{\mathop{88.2}\limits_{(+3.3)}}$   & $\boldsymbol{\mathop{91.3}\limits_{(+2.3)}}$  & $\boldsymbol{\mathop{88.9}\limits_{(+5.9)}}$     &  $\boldsymbol{\mathop{90.2}\limits_{(+4.7)}}$      &  $\boldsymbol{\mathop{84.5}\limits_{(+6.4)}}$    & $\boldsymbol{\mathop{96.8}\limits_{(+0.8)}}$     &  $\boldsymbol{\mathop{91.9}\limits_{(+1.0)}}$      &  $\boldsymbol{\mathop{91.3}\limits_{(+0.6)}}$  & $\mathop{97.9}\limits_{\underline{(+0.3)}}$  &  $\boldsymbol{\mathop{88.6}\limits_{(+0.2)}}$  &  $\boldsymbol{\mathop{90.2}\limits_{(+0.4)}}$   \\ \hline
\end{tabular}
\end{table*}

\begin{figure*}[t]
\center
\includegraphics[width=\textwidth]{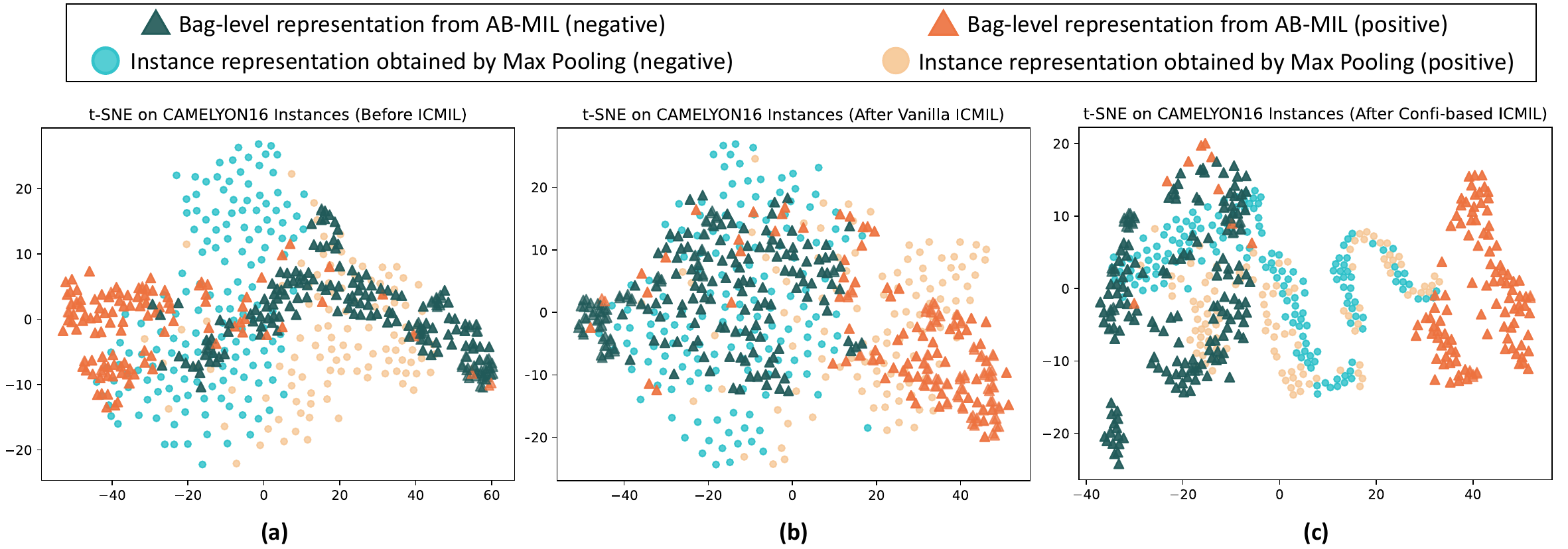}
\caption{The t-SNE visualization results of bag-level representations from ABMIL and the instances sampled with Max Pooling on each bag. (a) Results before ICMIL. (b) Results after 1 iteration of vanilla teacher-student ICMIL. (c) Results after 1 iteration of confidence-based teacher-student ICMIL.} 
% \vspace{-3mm}
\label{visualization}
\end{figure*}

\begin{figure*}[t]
\center
\includegraphics[width=\textwidth]{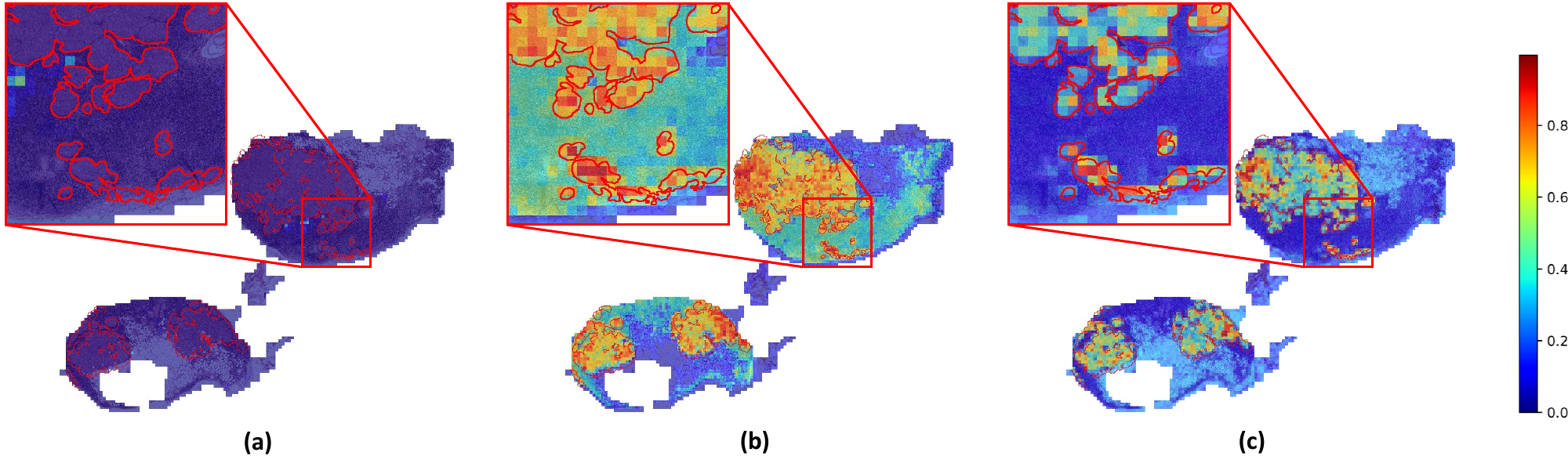}
\caption{Visualization of the attention scores of the instances in a bag before and after ICMIL. Tested with ABMIL backbone on Camelyon16. The areas surrounded by the red lines are the ground truth tumor areas, while the colored masks of the patches indicate their corresponding attention scores. (a) The baseline ABMIL method, which tends to only highlight very few instances.  (b) The vanilla ICMIL method, which can highlight the positive instance but struggles to mask the negative instances correctly. (c) The proposed confidence-based ICMIL method, which can not only highlight the positive instances but also correctly mask the negative instances.} 
% \vspace{-3mm}
\label{attention_visualization}
\end{figure*}

\subsection{Experimental Results}

% \begin{table}[t]
% \renewcommand\tabcolsep{4pt}
% % \renewcommand\arraystretch{1.3}
% \caption{Comparision between reweighting and resampling methods on confidence-based embedder fine-tuning. }
% \center
% \begin{tabular}{cc|cccc}
% \toprule
% Backbone                  & Method  & AUC ($\pm$STD)  & F1 ($\pm$STD)    & Acc ($\pm$STD) \\ \midrule
% \multirow{2}{*}{ABMIL}    & Reweighting & 91.8 ($\pm$0.35) & 82.5 ($\pm$0.41) & 86.6 ($\pm$0.42) \\
%                           & Resampling  & 92.5 ($\pm$0.89) & 86.0 ($\pm$1.76) & 90.6 ($\pm$1.07) \\ \midrule
% \multirow{2}{*}{DTFD-MIL} & Reweighting & 94.7 ($\pm$0.44) & 90.5 ($\pm$1.53) & 92.9 ($\pm$1.11) \\
%                           & Resampling  & 93.9 ($\pm$0.71) & 88.1 ($\pm$1.32)  & 90.9 ($\pm$0.80) \\ \bottomrule
% \end{tabular}
% \label{reweighting_resampling_comparison}
% \end{table}

\subsubsection{Ablation Study}
To verify the effectiveness of the teacher-student fine-tuning method, we conducted an ablation study comparing different designs. Our comparison includes the naive pseudo-label-based method, the vanilla teacher-student method, and the proposed confidence-based teacher-student method. Specifically, the naive pseudo-label method refers to the approach that directly uses the bag classifier to generate pseudo-labels for all instances and then fine-tunes the embedder based on the generated instance dataset. The vanilla teacher-student method and the confidence-based method are described in Section~\ref{vanillafinetuning} and Section~\ref{confidencefinetuning}, respectively.
% In general, the naive method and vanilla teacher-student method exhibit better generalization ability, making them compatible with most existing MIL backbones. On the other hand, the confidence-based method further leverages the aggregator trained in Classifier Phase to generate a confidence score for each instance, which often leads to higher performance. 
% Nevertheless, since the confidence-based fine-tuning relies on the aggregator to generate the confidence scores for the instances, this method will be equivalent to vanilla teacher-student fine-tuning method when used with Max-Pooling based MIL and Mean-Pooling based MIL, as the confidence score will be a constant for all the instances in a bag.

The experimental results are presented in Table~\ref{tab_ablation}. From the table, we observe that although the naive pseudo-label method provides some improvement to the MIL pipeline by enabling domain-specific embedder fine-tuning, it fails to surpass the performance boost achieved by the other two teacher-student-based methods. This is because directly using pseudo-labels generated by the fixed embedder and classifier for fine-tuning the pipeline itself does not introduce enough additional information. In contrast, both teacher-student-based fine-tuning methods lead to more satisfactory results across all three backbones. Specifically, the confidence-based teacher-student method usually outperforms the vanilla method as it incorporates the attention scores from the MIL aggregator for fine-grained weighting. By incorporating the attention scores into the pipeline, more valuable information is highlighted during the fine-tuning, resulting in more effective learning.

Nevertheless, it should be noted that the confidence-based fine-tuning method can degrade to vanilla fine-tuning method for Max Pooling-based and Mean Pooling-based MIL methods, as they use non-learnable pooling layers as their aggregators, which generate one-hot or constant attention scores for all the instances. Using one-hot attention scores as the input for the converting layer will result in a constant 1 confidence score output, which is equivalent to not using the confidence-based scheme. Using constant attention scores (i.e., $\frac{1}{K}$) will lead to constant confidence scores. A constant confidence score for the instances will not affect the relative distribution of the training samples, making this confidence-based method still theoretically equivalent to the vanilla fine-tuning method.
\subsubsection{Parameter Study about the Converting Layer}
We further conduct a parameter study on $\beta$ of the converting layer. $\beta$ controls the number of uncertain instances that need to be masked during the confidence-based fine-tuning. As shown in TABLE~\ref{parameter_study}, $\beta$=6 leads to the highest performance on ABMIL, while $\beta$=2 brings about the best results for DTFD-MIL. Generally speaking, the performance of ICMIL is not very sensitive to this parameter as long as it remains in a reasonable range. We empirically recommend using $\beta$=6 for most circumstances. 

\subsubsection{Comparison of Different Augmentation Methods}

We have investigated different bag augmentation methods based on the three MIL backbones in our experiments, and the experimental results are presented in TABLE~\ref{augmentation_comparison}. From the table, we can learn that bag augmentation can bring performance improvement under most circumstances. When comparing the sole bag mix-up augmentation with the mix-up + pseudo-bag scheme, we can learn that the latter can usually outperform the former. This proves that the mix-up + pseudo-bag augmentation method can bring more diversity to the dataset, leading to better generalization ability of the model. For example, for Max Pooling-based MIL and ABMIL, mix-up + pseudo-bag augmentation lead to a remarkable boost in all three metrics on the Camelyon16 and HCC datasets. 
% As to the TCGA-Lung dataset and TCGA-RCC dataset, the improvement is comparatively moderate because the baseline methods have already reached a very high performance, leaving very limited space for improvement. 

% In contrast, it is also shown that the augmentation method leads to relatively small improvement when used with the DTFD-MIL backbone. This is mainly because the DTFD-MIL is a two-tier pseudo-bag-based MIL method, whose pipeline already involves the pseudo-bag augmentation process at the beginning. Therefore, repeatedly applying our bag augmentation method can only bring limited extra diversity to the training samples, resulting in less performance increase.

% Additionally, 

\subsubsection{Study on the Ceasing Point of Iteration}

As an iterative optimization algorithm, determining the proper ceasing point for the iteration is vital. Therefore, we present an ablation study on Camelyon16 to find the most cost-effective creasing point for ICMIL. Each iteration involves one million patches for Embedder Phase. The experimental results of the performance based on different training iterations are presented in Fig.~\ref{ablation_study_iteration}. From the graph, we can learn that the general performance of all three backbones tends to grow at the early iterations, and then become steady for the next few iterations. Although in certain cases some MIL backbones may achieve the highest AUC when trained with 2 ICMIL iterations, the extra performance boost is minor, especially when considering the extra training time. Therefore, we empirically recommend running ICMIL for one iteration.

\subsubsection{Comparison with Other Methods}

We also have a comprehensive comparison between existing MIL classification methods and ICMIL. The experimental results are presented in Table~\ref{experimental_results}. As shown, confidence-based ICMIL that incorporates bag-level augmentation brings consistent improvements to the three backbones on all the datasets. Taking Camelyon16 as an example, ICMIL improves the AUC of Max Pooling-based MIL by 7.9\% and ABMIL by 7.0\%. When used with the current state-of-the-art MIL model DTFD-MIL, ICMIL further increases its performance by 1.8\% AUC, 3.3\% F1, and 2.3\% Acc, demonstrating its effectiveness. 

For the HCC dataset, the performance boost is also very obvious. Since risk grading is a more difficult task, the quality of the instance-level representations plays a crucial role in generating more separable bag-level representations. Therefore, after applying ICMIL, the three MIL backbones all gain great performance surge on the HCC dataset.

As to the two TCGA datasets, it is shown that the results of all the baseline MIL models are very high. Among them, Max Pooling-based MIL, ABMIL, and DTFD-MIL all achieved an AUC over 95\%, leaving very little room for improvement. Nonetheless, ICMIL still bring consistent improvements to all three MIL backbones. On TCGA-Lung, for the state-of-the-art DTFD-MIL, ICMIL pushes its performance even further, leading to 96.8\% AUC, 91.9\% F1, and 91.3\% Acc; On TCGA-RCC, ICMIL improves the performance of DTFD-MIL to 97.9\% AUC, 88.6\% F1, and 90.2\% Acc. 

\subsubsection{Visualization Analysis}

In Fig.~\ref{visualization}, we present an intuitive comparison of the representations before and after ICMIL. Fig.~\ref{visualization}(a) is the t-SNE visualization of the bag representations generated by ABMIL before ICMIL, with some instances sampled by Max Pooling from the bags in the background. It is shown that both the bag-level and the instance-level representations are somehow separable by two different boundaries. Fig.~\ref{visualization}(b) presents the visualization results after 1 iteration of vanilla teacher-student embedder fine-tuning. As shown, the separability of the bag-level representations is improved. Besides, the bag representations also become more aligned with the instance representations, which is the natural result when using the bag-level decision boundaries to guide the instance-level embedder fine-tuning. However, when we change to use the proposed confidence-based embedder fine-tuning method, such alignment characteristics are weakened. This is because in this case, the fine-tuned representations from the embedder tend to be more tailored for the latent space of ABMIL, making the instance representations sampled by max pooling become less separable as shown in Fig.~\ref{visualization}(c).

Additionally, in Fig.~\ref{attention_visualization}, we present a visual comparison of the highlighted instances in a bag before and after ICMIL. In Fig.~\ref{attention_visualization}(a), it is illustrated that the baseline ABMIL method tends to generate high attention scores only for a few distinctive instances. This is because it uses a frozen patch embedder and cannot project the WSI instances to an ideal subspace, in which case the aggregator tends to find the one single instance that can represent the entire bag. In comparison, as shown in Fig.~\ref{attention_visualization}(b), most of the tumor area patches are correctly highlighted after ICMIL. This indicates that the fine-tuned embedder finds a better latent space to project the positive and negative instances. However, it is also shown that many negative patches around the real tumor areas still receive incorrect high attention scores. In contrast, the proposed confidence-based ICMIL shows better performance as shown in Fig.~\ref{attention_visualization}(c). It can not only highlight the positive instances correctly, but also dim the negative patches properly. This proves the effectiveness of the confidence-based ICMIL.

\section{Conclusion and Discussion}
\label{section_conclusion}
In this study, we have proposed the idea that a bag-level classifier can be a good instance-level teacher, and introduced ICMIL, a simple yet powerful framework for WSI classification. The primary objective of ICMIL is to bridge the gap between the patch embedding stage and the bag-level classification stage in existing MIL backbones. To address this challenge at a low cost, ICMIL proposes to couple the two stages in an iterative manner. To achieve an effective embedder fine-tuning process in Embedder Phase, we further propose a confidence-based teacher-student framework to facilitate category knowledge transfer from the bag classifier to the patch embedder. The extensive experiments on four different datasets using three representative MIL backbones demonstrate that ICMIL consistently enhances the performance of existing MIL backbones, achieving state-of-the-art results. 

The limitations of the proposed confidence-based ICMIL mainly lie in that the converting layer used during the confidence score generation process is very simple and in certain cases it may require careful parameter tuning to avoid masking too many instances of uncertainty. In our future works, we will investigate and compare more different converting functions for the Embedder Phase and try to propose a self-adaptive converting layer for more accurate confidence score generation.

% The limitations of the proposed confidence-based ICMIL mainly lie in the converting layer. The current converting layer is still very simple and in certain cases may require careful parameter tuning to avoid masking too many uncertain instances. In our future works, we will investigate and compare more different converting functions for Embedder Phase and try to propose a self-adaptive converting layer for more accurate confidence score generation.

% In this work, we propose a simple yet effective WSI classification framework named ICMIL. ICMIL focuses on bridging the gap between the patch feature extraction stage and the bag-level classification stage in existing MIL backbones. To address this challenge with low computational cost, ICMIL proposes to couple the two stages in an iterative manner. Classifier Phase of ICMIL fixes the embedder and trains the MIL classifier on the bag-level, while Embedder Phase turns to fixing the MIL classifier and fine-tuning the patch embedder on the instance-level. To realize an effective and efficient embedder fine-tuning process for Embedder Phase, we also propose a teacher-student-based framework, which can be further extended into the vanilla teacher-student fine-tuning method and confidence-based teacher-student fine-tuning method in this work. We have tested ICMIL on three different datasets with three representative MIL backbones, and the experimental results show that ICMIL can bring a consistent performance boost to existing MIL backbones, achieving new state-of-the-art results.

% \section*{Acknowledgments}

% None

\bibliographystyle{IEEEtran.bst}
\bibliography{IEEEabrv.bib}

\end{document}